\documentclass[runningheads]{llncs}
\usepackage{graphicx}
\usepackage{amsmath}
\begin{document}
\title{No Adjective Ordering Mystery, and No Raven Paradox, Just an Ontological Mishap}
%
%
\author{Walid S. Saba}
\authorrunning{W. Saba}
\titlerunning{Just an Ontologicaal Mishap}
%
\institute{Astound.ai, 111 Independence Drive, Menlo Park, CA 94025 USA \\
\email{walid@astoun.ai}}
\maketitle              
\section{Where Logical Semantics (Might Have) Went Wrong}
In the concluding remarks of \textit{Ontological Promiscuity} Hobbs (1985) made what we believe to be a very insightful observation: given that semantics is an attempt at specifying the relation between language and the world, if ``”one can assume a theory of the world that is isomorphic to the way we talk about it ... then semantics becomes nearly trivial''”. But how exactly can we rectify our logical formalisms so that semantics, an endeavor that has occupied the most penetrating minds for over two centuries, can become (nearly) trivial, and what exactly does it mean to assume a theory of the world in our semantics?

In this paper we hope to provide answers for both questions. First, we believe that a commonsense theory of the world can (and should) be embedded in our semantic formalisms resulting in a logical semantics grounded in commonsense metaphysics. Moreover, we believe
the first step to accomplishing this vision is rectifying what we think was a crucial oversight in logical semantics, namely the failure to distinguish between two fundamentally different types of concepts: (\textit{i}) \textbf{ontological concepts}, that correspond to what Cocchiarella (2001) calls first-intension concepts and are types in a strongly-typed ontology; and (\textit{ii}) \textbf{logical concepts} (or second intension concepts), that are predicates corresponding to properties of (and relations between) objects of various ontological types\footnote{Our ontological types are similar to the types in Sommers' (1963) \textit{Tree of Language}}.

In such a framework,  which we will refer to henceforth by \textsc{ontologik}, it will be shown how type unification and other type operations can be used to account for the `missing text phenomenon' (MTP) (see Saba, 2019a) that is at the heart of most challenges in the semantics of natural language, by ’uncovering’ the significant amount of missing text that is never explicitly stated in everyday discourse, but is often implicitly assumed as shared background knowledge.

\section{The `Missing Text Phenomenon' (MTP)}
To breifly describe the motivation behind our work, let us consider a simple example that illustrates the relationship between MTP and the need for distinguishing between ontological and logical concepts. Consider the sentences in (1) and (2) and their translation into first-order
predicate logic (FOPL).
\newline
(1) \hspace{0.5 mm} \emph{Julie is an articulate person}
\newline \indent \indent  $\Rightarrow$ {\textbf{articulate}}(\emph{Julie}) $\land$ {\textbf{person}}(\emph{Julie})
\newline
(2) \hspace{0.5 mm} \emph{Julie is articulate}
\newline \indent \indent $\Rightarrow$ {\textbf{articulate}}(\emph{Julie})
\newline
\newline
Note that (1) and (2) have different translations into first-order predicate logic, although the two sentences seem to have the same semantic and cognitive content. For the translation of (1) and (2) into FOPL to be the same, {\textbf{person}}(\emph{Julie}) in (1) must be assumed to be true, \textit{a priori} since (\emph{a} $\land$ \emph{b} = \emph{a}) $\supset$ \emph{b}. This in fact is quite sensible, since {\textbf{articulate}}(\emph{Julie}) is meaningful only if \emph{Julie} is a person. What all of this means is that the proper translation of (1) and (2) should treat {\textbf{person}} as a type, and {\textbf{articulate}} as a property that is predicable (makes sense to say) of objects that are of type person:
\newline
\newline
(3) \hspace{0.5 mm} ($\exists$$^{1}$\emph{Julie} :: \texttt{person})(\textbf{articulate}(\emph{Julie}))
\newline
\newline
What (3) says is that there is a unique object named \emph{Julie}, such that the property \textbf{articulate} is true of \emph{Julie}. The point of this simple example was to illustrate that \texttt{person} in (2) was implicit, and was assumed by the speaker to be `recoverable' given our commonsense knowledge of the world. Embedding ontological types in our semantics thus allows us to uncover all the missing text and, in the process, resolve some of the most challenging problems in the semantics of natural language. Consider (4) and its translation into a logical form in \textsc{ontologik}.
\newline
\newline
(4) \emph{The loud omelet wants another beer}
\newline \indent  $\Rightarrow$  ($\exists$\emph{o} :: \texttt{omelet})($\exists$\emph{b} :: \texttt{beer}) 
\newline
 \indent \indent  \indent  (\textbf{loud}(\emph{o} :: \texttt{person}) $\land$  \textbf{want}(\emph{o} :: \texttt{animal}, \emph{b} :: \texttt{entity}))
\newline
\newline
That is, there is some \emph{o}, which is an object of type \texttt{omelet}, and some \emph{b}, which is an object of type \texttt{beer}, and such that \emph{o} is \textbf{loud}, and thus be must (in this context) an object of type \texttt{person}, and where \emph{o} wants \emph{b}, and where in this context \emph{o} must be an object of type \texttt{animal} and where \emph{b} could be any \texttt{entity}. What we have now are objects that, in the same scope, are associated with different types. The type unification (\texttt{beer} $\bullet$ \texttt{entity}) is quite simple, since \texttt{beer} \emph{IsA} \texttt{entity}. However, consider now the type unification concerning \emph{o} that allows us to `ucover' the missing text, namley that [\textbf{some loud person eating}] \emph{the omelet wants a beer}:
\newline \newline
(\texttt{omelet}  $\bullet$ (\texttt{animal}  $\bullet$  \texttt{person}))  $\rightarrow‎$ (\texttt{omelet}  $\bullet$ \texttt{person}) 
\newline $\rightarrow‎$ (\texttt{person} \textsc{eating} \texttt{omelet}) 
\section{What Adjective-Ordering Restrictions Mystery?}
The phenomenon of adjective-ordering re-strictions (AORs) concerns the apparent adjective ordering we tend to prefer when multiple adjectives are used in a sequence. The AOR phenomenon, which has been a longstanding subject of debate in linguistics (see Cinque 1994; and Vendler, 1968), can be illsutrated by the example in (5), where the first sentence is usually preferred.
\newline
\newline
(5) \hspace{0.5 mm} \emph{Jon bought a beautiful red car}
\newline
\indent  \hspace{0.5 mm} \#\emph {Jon bought a red beautiful car}
\newline
\newline
Far from being a `mystery', however, it turns out that the rules of type-casting are behind our decision of a preferred reading. Assuming \textbf{beautiful}(\emph{x} :: \texttt{entity}) and \textbf{red}(\emph{x} :: \texttt{physical}) - i.e., that \textbf{beautiful} is predicable of any \texttt{entity} and that \textbf{red} is predicable of objects that must be of type \texttt{physical}, then the type unification (\texttt{entity}  $\bullet$ (\texttt{physical}  $\bullet$  \texttt{car}))  will succeed while (\texttt{physical}  $\bullet$ (\texttt{entity}  $\bullet$  \texttt{car})) is technically not allowed: we can always cast-up (generalize), casting down (from \texttt{entity} to \texttt{physical}, in this case), is however undecidable. 
\section{What Paradox of the Ravens?}
Introduced in the 1940‟s by (Hempel, 1945) the paradox of the raven can be stated as follows: H1 is logically equivalent to H2, and thus both hypotheses should be equally confirmed (or disconfirmed) by the same observations. 
\newline\newline
(H1) \hspace{0.5 mm} \emph{All ravens are black} $\Rightarrow$ ($\forall$\emph{x})(\textbf{raven}(\emph{x}) $\supset$ \textbf{black}(\emph{x}))\\
(H2) \hspace{0.5 mm} \emph{All non-black things are non-ravens} $\Rightarrow$ ($\forall$\emph{x})($\neg$\textbf{black}(\emph{x}) $\supset$ $\neg$\textbf{raven}(\emph{x}))\
\newline\newline
However, that leaves us with the un-pleasant conclusion that observing a red ball, or blue shoes, both of which confirm H2, also confirm that all ravens are black, which clearly could not be accepted. Again, far from being a paradox, we argue that the problem lies in our FOPL representation of H1 and H2, namely the universal use of predication to represent genuine predicates as well as ontological types. Instead, we argue that both H1 and H2 have one and the representation, namely ($\forall$\emph{x} :: \texttt{raven})(\textbf{black}(\emph{x})) - that is, \textbf{black} is true of every object of type \texttt{raven}, which is equally confirmed and disconfirmed by the same observations (due to space limitations we cannot here show how in \textsc{ontologik} H1 and H2 would get the same logical form, but interested readers can see (Saba, 2019b)).
\newline \newline
\textbf{References}
\newline \newline
Cocchiarella, N. (2001), Logic and Ontology, \emph{Axiomathes}, 12 (1-2), pp. 117-150. \\
Hobbs, J. (1985), Ontological Promiscuity, \emph{Proceedings of ACL-1985}, pp. 60-69. \\
Hempel, C. G. (1945), Studies in the Logic of Confirmation, \emph{Mind}, 54, pp. 1-26. \\
Saba,W. S. (2019a), On the Winograd Schema, \emph{FLAIRS-2019}, AAAI Press. \\
Saba,W. S. (2019b), Language and its commonsense: where logical semantics \\ \indent went wrong, and where it can (and should) go, \texttt{https://bit.ly/2PcqltM}. \\
Sommers, F. (1963),  Types and Ontology, \emph{ Philosophical Review}, 72, pp. 327-363.

\end{document}